\documentclass[conference,10pt]{IEEEtran}
\IEEEoverridecommandlockouts

\usepackage[utf8]{inputenc}
\usepackage{cite}
\usepackage{amsmath,amssymb}
\usepackage{graphicx}
\usepackage{booktabs}
\usepackage{multirow}
\usepackage{balance}
\usepackage{caption,subcaption}
\usepackage{url}

\title{Edge-Based Standing-Water Detection via FSM-Guided Tiering and Multi-Model Consensus}

\author{
    \IEEEauthorblockN{Oliver Larsen$^{1}$, Mahyar T. Moghaddam$^{1}$}
    \IEEEauthorblockA{$^{1}$SDU – University of Southern Denmark, MMMI \\
    Odense, Denmark \\
    Email: \{olar, mtmo\}@mmmi.sdu.dk}
}

\usepackage{tikz}
\usepackage{textcomp}

\begin{document}
\maketitle
\footnotetext{\textcopyright~2026 IEEE. Personal use of this material is permitted. Permission from IEEE must be obtained for all other uses, in any current or future media, including reprinting/republishing this material for advertising or promotional purposes, creating new collective works, for resale or redistribution to servers or lists, or reuse of any copyrighted component of this work in other works. Accepted at the In Practice Track of IEEE ICSA 2026.}

\begin{abstract}
Standing water in agricultural fields threatens vehicle mobility and crop health. This paper presents a deployed edge architecture for standing-water detection using Raspberry-Pi-class devices with optional Jetson acceleration. Camera input and environmental sensors (humidity, pressure, temperature) are combined in a finite-state machine (FSM) that acts as the architectural decision engine. The FSM-guided control plane selects between local and offloaded inference tiers, trading accuracy, latency, and energy under intermittent connectivity and motion-dependent compute budgets. A multi-model YOLO ensemble provides image scores, while diurnal-baseline sensor fusion adjusts caution using environmental anomalies. All decisions are logged per frame, enabling bit-identical hardware-in-the-loop replays. Across ten configurations and sensor variants on identical field sequences with frame-level ground truth, we show that the combination of adaptive tiering, multi-model consensus, and diurnal sensor fusion improves flood-detection performance over static local baselines, uses less energy than a naive always-heavy offload policy, and maintains bounded tail latency in a real agricultural setting.
\end{abstract}


\section{Introduction}
\label{sec:introduction}
Standing water in agricultural fields threatens vehicle mobility and crop health, especially on narrow off-road tracks where drivers must make rapid decisions under changing light, weather, and motion~\cite{silva2024edge,islam2022off}. In such settings, detection must run on commodity edge hardware mounted in the vehicle, with intermittent connectivity, constrained power, and strict tail-latency requirements~\cite{overview_edge_agri_iot_2020}. A single heavy vision model or cloud-based inference is often infeasible~\cite{cloud_edge_agri_2025}; the system must combine noisy camera and sensor inputs while explicitly managing latency, energy, and caution.
This paper presents a deployed edge architecture for standing-water detection on off-road vehicles using Raspberry-Pi-class devices with an optional Jetson accelerator. Camera frames and environmental sensor readings are ingested on a \emph{Gathering} node, while a \emph{Processing} node owns the finite state machine (FSM), diurnal baselines, and configuration. The FSM selects inference tiers and offloading policies based on scores, motion, and resource state; a multi-model YOLO ensemble produces image scores; and diurnal sensor fusion adjusts these scores using humidity, temperature, and pressure anomalies. All intermediate values and decisions are logged per frame, enabling bit-identical replays and controlled ablations on the production hardware. The system is developed for a real agricultural use case within a Digital Research Centre Denmark (DIREC)-funded project, in collaboration with the DIREC industry consortium on agriculture.
The work is structured around three hypotheses:
\begin{itemize}
\item \textbf{H1 (efficiency from adaptive tiering):} FSM-guided, motion-aware tiering reduces unnecessary heavy-tier use and total energy relative to naive heavy-model policies, while maintaining or improving flood-detection quality.
\item \textbf{H2 (accuracy from multi-model consensus):} Multi-model consensus improves binary macro F1 and balanced accuracy compared to single-model variants at similar energy cost, with higher tail latency as the main trade-off.
\item \textbf{H3 (caution from diurnal sensor fusion):} Diurnal sensor fusion adjusts system caution and temporal coverage across environmental regimes (flood-like vs.\ hot, dry anomalies) without substantially degrading flood recall.
\end{itemize}
Contributions are: \textbf{C1:} Multi-view description of a three-node architecture (Gathering Pi, Processing Pi, Jetson Worker) with single-state owner and schema-stable messages; \textbf{C2:} Unified control plane integrating FSM tiering, YOLO consensus, and sensor fusion with per-frame logs; \textbf{C3:} Hardware-in-the-loop evaluation across ten ablations and sensor variants quantifying accuracy, energy, coverage, and latency effects.
Section~\ref{sec:related} reviews related work. Section~\ref{sec:architecture} presents the architecture and Section~\ref{sec:decision} details decision logic. Section~\ref{sec:evaluation} provides experiments, evaluates H1--H3, and discusses implications and limitations. Section~\ref{sec:conclusion} concludes the paper.

\section{Background and Related Work}
\label{sec:related}

This section briefly situates our system within work on standing-water detection and edge perception architectures, and motivates our FSM-guided, ensemble-based design with diurnal sensor fusion.

\subsection{Standing-Water Detection and Off-Road Perception}

Standing-water or flood-hazard detection for off-road vehicles aims to alert a vehicle’s control system to hazardous ponds or flooded fields in real time. Prior work spans remote sensing, fixed camera networks, and on-vehicle perception. Satellite and airborne remote sensing provide wide-area flood maps at low cost using optical, infrared, and microwave sensors, but temporal and spatial resolution limits and cloud cover reduce their usefulness for fine-grained, real-time perception along specific tracks~\cite{tao2024review}. Ground-level datasets such as AlleyFloodNet were introduced because many flood-detection studies relied on aerial imagery; they highlight that datasets targeting economically vulnerable, flood-prone areas and localised scenes remain scarce~\cite{lee2025alleyfloodnet}.

To monitor rivers or roads, surveillance-camera systems segment water regions using deep networks and may infer water levels from segmentation masks or landmarks~\cite{lee2024dnnflood, jan2022realtime}. These GPU-based systems assume stable power, daylight data, and more capable hardware than Raspberry-Pi-class boards, even when all processing is pushed to the edge~\cite{tao2024review, jan2022realtime}.

Image-based flood monitoring has also adopted multimodal sensor fusion: RGB and long-wave infrared (LWIR) cameras can provide robust flood detection day and night, and network-slimming techniques enable deployment on embedded IoT edge devices~\cite{lee2024dnnflood}. Across these lines of work, systems generally assume stable connectivity, off-board processing, or more powerful hardware than commodity Raspberry-Pi-class devices. There is little work on real-time on-vehicle flood detection operating on such hardware under strict latency and energy constraints~\cite{tao2024review, jan2022realtime}.

\subsection{Resource-Aware Edge Architectures and Perception Tactics}

Edge computing architectures bring computation closer to data sources, reducing latency and network congestion~\cite{dauda2024survey}. The Thing-to-Thing Research Group’s definition of edge computing emphasises that substantial processing and storage resources are placed near sensors and actuators; multi-tiered fog architectures distribute compute layers between end devices and the cloud~\cite{hong2024iot}. Edge systems are commonly organised into sensor or camera hubs connected to gateway nodes via message-broker protocols such as MQTT or Kafka, with gateways aggregating data and forwarding it to local servers or the cloud. Surveys of IoT architectures note that most prior work concentrates on throughput and latency optimisation and on dynamic offloading strategies~\cite{dauda2024survey, hong2024iot}, but rarely includes an explicit, stateful control plane that can orchestrate perception tiers or maintain deterministic, per-frame logs needed for replayable evaluation on resource-constrained hardware.

Perception tactics for resource-aware inference include cascaded or tiered models, ensembles, and sensor fusion. Adaptive schemes run lightweight models first and forward only ambiguous samples to more powerful models, often using confidence or latency budgets to trigger escalation or offloading~\cite{kolawole2024}. These cascades reduce average inference cost and can distribute models across tiers, for example small models at the edge and larger ones in the cloud~\cite{kolawole2024}. In flood monitoring, multi-sensor fusion (e.g., RGB with LWIR, or multispectral with SAR) improves detection robustness under challenging conditions~\cite{lee2024dnnflood, dauda2024survey}. Nonetheless, ensembles and sensor fusion are typically treated as model-level choices rather than first-class architectural elements.

\paragraph*{Positioning of This Work} In contrast, our system makes these tactics explicit at the architectural level. We employ an FSM-controlled control plane that selects perception tiers based on confidence, motion cues, and resource state; a three-model YOLO ensemble with deterministic consensus scoring; and diurnal sensor fusion that uses environmental priors (e.g., day/night and humidity/temperature anomalies) to modulate caution on Raspberry-Pi-class edge devices. These tactics are evaluated through hypotheses H1–H3 on real hardware in Section~\ref{sec:evaluation}.

\section{Architecture}
\label{sec:architecture}
This section provides a multi-view architectural description of the flood-hazard detection system. We first introduce the logical responsibilities of the main subsystems, then describe their deployment on concrete hardware, and finally outline the runtime behaviour from ingestion to decision persistence.

\begin{figure}[t]
\centering
\includegraphics[width=0.8\columnwidth]{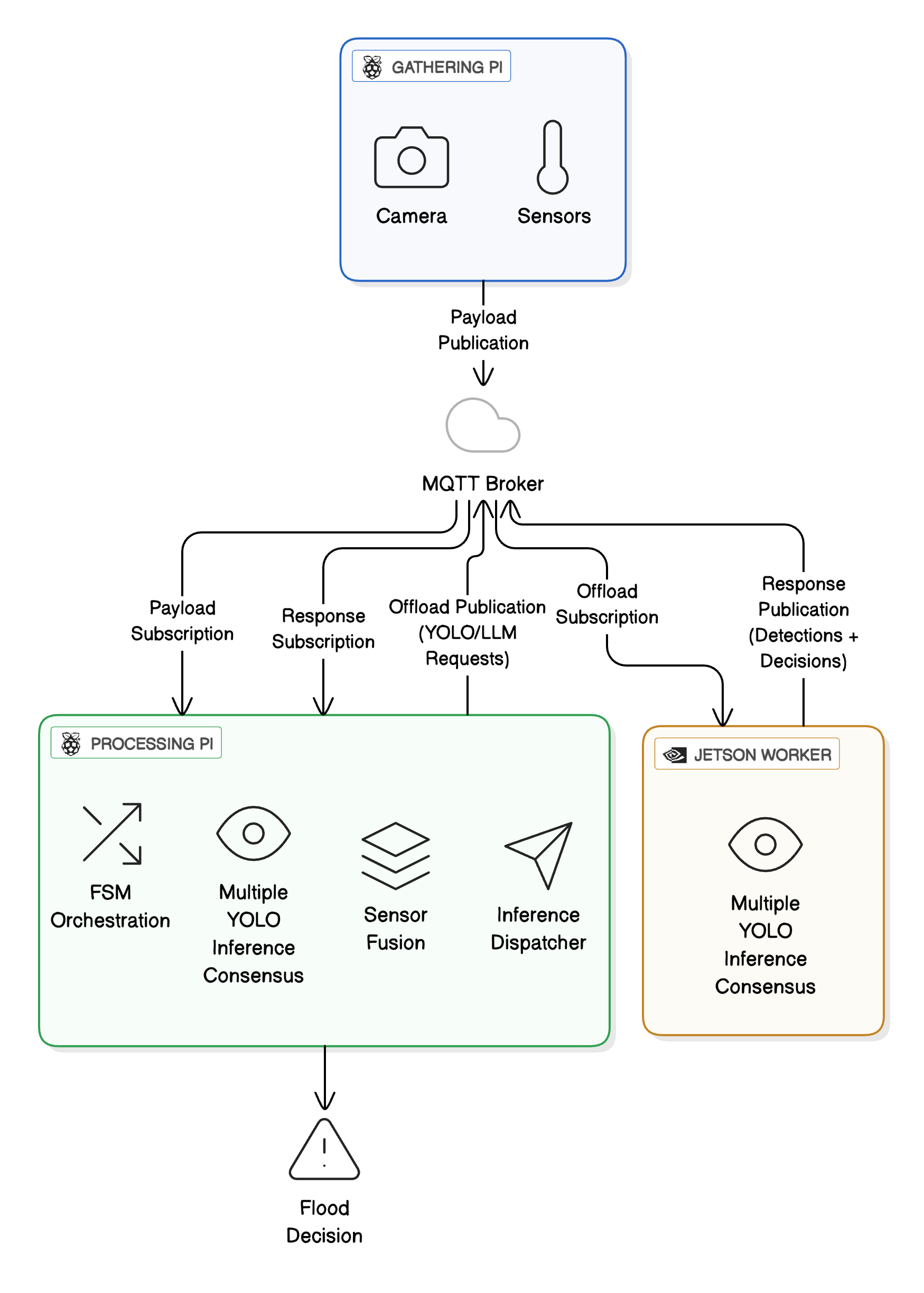}
\caption{High-level logical view of the three-node flood-detection architecture}
\label{fig:high_level}
\end{figure}

\subsection{Logical View}

At a logical level, the system consists of three cooperating edge nodes with clearly separated responsibilities and a single owner of architectural state. The focus of this view is on the roles of each subsystem, the partitioning of long-lived state, and the high-level data flows shown in Figure~\ref{fig:high_level}.

\textbf{Gathering Pi.}
The Gathering Pi is a pure data-acquisition component. It captures RGB frames together with temperature, humidity, and pressure readings, enriches each frame with timestamp, motion label, and basic metadata, and emits a single, schema-stable message type. This unified stream abstracts away sensor heterogeneity and forms the sole ingress into the rest of the system.

\textbf{Processing Pi.}
The Processing Pi acts as the architectural ``brain'' of the system and is the only node that maintains long-lived architectural state, including the FSM, diurnal baselines, and configuration parameters. Logically, it provides four processing capabilities:
\begin{itemize}
    \item \emph{Baseline and anomaly computation}, which derives environmental anomalies from diurnal baselines;
    \item \emph{FSM-guided orchestration}, which determines model tier selection and routing policy;
    \item \emph{Inference and consensus}, which combines outputs from local or offloaded YOLO ensembles into a unified image score; and
    \item \emph{Scoring and classification}, which produces per-frame hazard labels and enriches them with metadata.
\end{itemize}
The Processing Pi is the authoritative source of system decisions and the only component that updates architectural state or produces final hazard outputs.

\textbf{Jetson Worker.}
The Jetson Worker is a stateless accelerator that optionally executes heavy-tier YOLO inference. It accepts inference jobs for small, medium, and large model variants and returns aggregated detections and timing metadata.

\subsection{Deployment View}

\begin{figure*}[t]
\centering
\includegraphics[width=\textwidth]{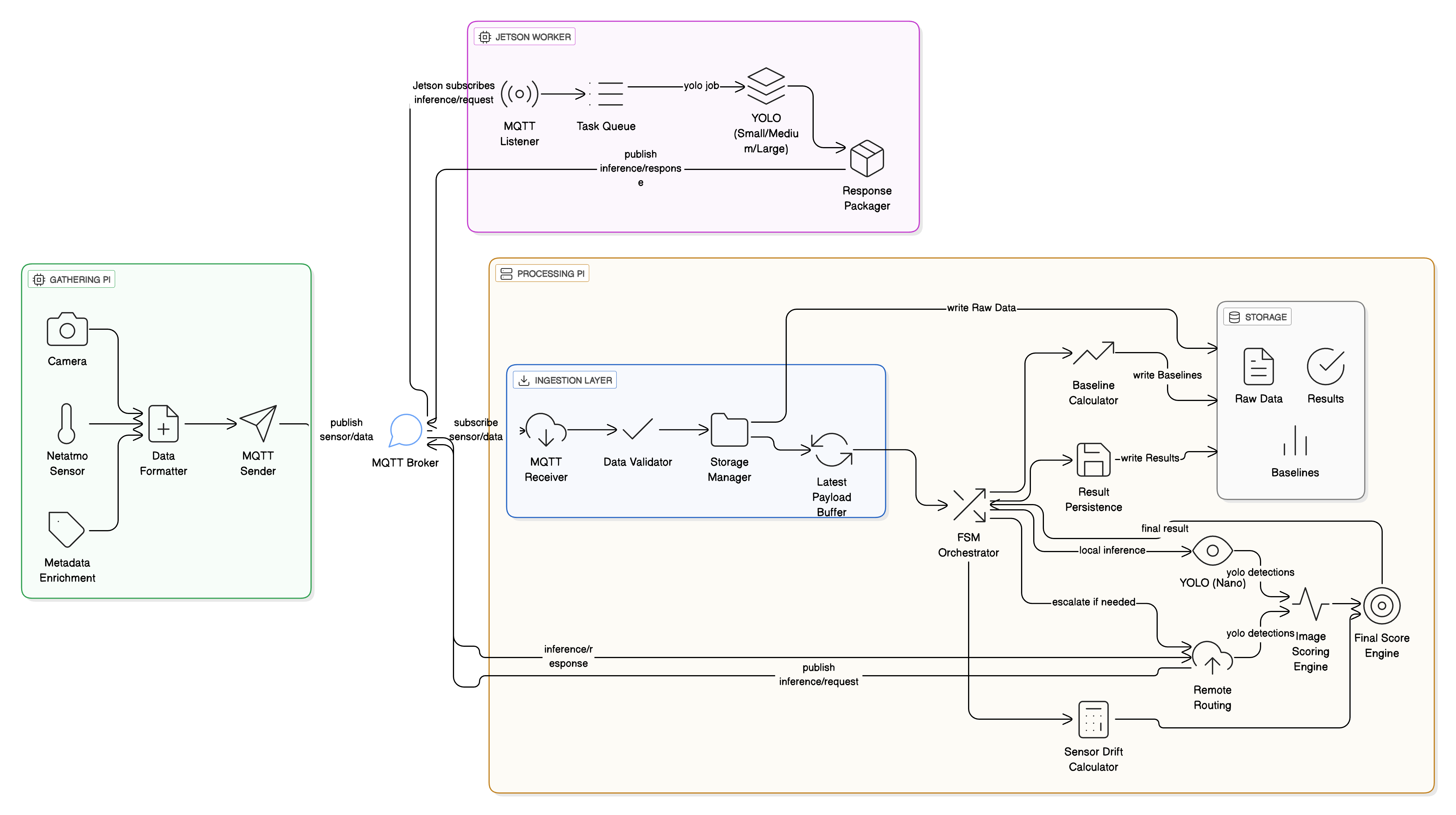}
\caption{Deployment view of the system.}
\label{fig:low_level}
\end{figure*}

The deployment view maps the logical responsibilities onto concrete hardware and communication channels, as shown in Figure~\ref{fig:low_level}.

\textbf{Nodes and network.}
The system runs on three physical devices mounted in the vehicle:

\begin{itemize}
    \item \emph{Gathering Pi (Pi-G)}: a Raspberry Pi-class board co-located with the camera and environmental sensors;
    \item \emph{Processing Pi (Pi-P)}: a Raspberry Pi-class board with slightly higher memory and I/O bandwidth, acting as the central orchestrator; and
    \item \emph{Jetson Worker (Jetson-J)}: an NVIDIA Jetson AGX Orin-class module providing GPU acceleration for heavy YOLO tiers.
\end{itemize}

All nodes communicate over an in-vehicle network using an MQTT broker. Topics are deliberately narrow and stable: \texttt{sensor/data} carries input frames and sensor readings; \texttt{inference/request} and \texttt{inference/response/\textless job\_id\textgreater} encapsulate offloaded inference jobs and results; and \texttt{inference/jetson/status} is a retained heartbeat that advertises Jetson health to the Processing Pi. This event-driven message bus decouples producers and consumers, enabling independent evolution of the Gathering, Processing, and Jetson services.

\subsection{Behavioural View}

The behavioural view describes the system's runtime operation, focusing on the per-frame decision workflow.

\textbf{Normal decision pipeline.}
For each message on \texttt{sensor/data}, the Processing Pi executes a fixed sequence:

\begin{enumerate}
    \item \emph{Ingestion and anomaly computation}: parse the payload and compute $\Delta T$, $\Delta \mathrm{RH}$, and $\Delta P$ relative to the current baselines.
    \item \emph{FSM decision}: use the current FSM state (S0--S4), motion label, and previous score to select the model tier and routing policy.
    \item \emph{Inference}: run the chosen YOLO ensemble either locally (nano) or via an offloaded job to the Jetson (small/medium/large), and aggregate detections into an image score.
    \item \emph{Scoring}: apply the sensor boost to obtain the combined score and classify it into $0 =$ No Flood, $1 =$ Some Water, $2 =$ Flooded.
    \item \emph{State update and logging}: update the FSM state using hysteresis and write a per-frame JSON record with all intermediate values.
\end{enumerate}

Pipeline instances may overlap (e.g., while waiting for a Jetson response), but all logic and state transitions are deterministic, ensuring reproducible outcomes for the same input stream.

\textbf{Offloading and fallback.}
When a heavy tier is selected, the Processing Pi offloads an inference job to the Jetson and waits for aggregated detections. Missing heartbeats or timeouts mark the Jetson as unavailable: the system falls back to local nano-tier inference and the FSM enters the resource-constrained state~S4 until Jetson health signals resume.

\subsection{Reliability and Operability Tactics}

The architecture uses a small set of tactics to keep behaviour predictable under resource constraints and intermittent connectivity.

\textbf{Bounded queues and backpressure.}
The Processing Pi keeps only the latest frame in its offload buffer, and the Jetson processes a single inference job at a time. This “latest-wins” design trades some temporal density for bounded latency, memory, and energy.

\textbf{Health checks and circuit breakers.}
A retained \texttt{inference/jetson/status} heartbeat and the \texttt{MQTT\_INFERENCE\_TIMEOUT} guard the offload path: repeated timeouts or missing heartbeats cause the Processing Pi to stop offloading, run local tiers only, and enter S4 until Jetson health stabilises.

\textbf{Reproducibility and observability.}
Per-frame JSON logs and stable message schemas make the runtime behaviour executable: every decision can be traced back to inputs, scores, and FSM state, and replay scripts can regenerate results from \texttt{storage/data/} into \texttt{storage/data\_results/} for controlled ablations.

\textbf{Security and privacy.}
All processing runs on-board the vehicle with no online dependency, and MQTT traffic stays within the in-vehicle network.

\section{Decision Logic and Orchestration}
\label{sec:decision}

This section describes how the FSM, tier-selection policy, consensus aggregation, and sensor fusion work together to produce resource-aware flood decisions.

\subsection{Motion-Aware Finite-State Machine}

The Processing Pi maintains a finite-state machine over five states:
$S0$ (\textbf{Normal Watch}), $S1$ (\textbf{Uncertainty Investigation}), $S2$ (\textbf{Confirmed Flood}), $S3$ (\textbf{Ambiguity/Conflict Resolution}), and $S4$ (\textbf{Resource-Constrained}). For each frame, the FSM takes as input the current state, the combined score $c$ (from the unified scoring pipeline), the motion cue $m$ (stopped/slow/fast), and a resource flag $r$ (normal/constrained), and outputs a next state $s'$ and a selected tier $t$.

The combined score is partitioned into three bands by two thresholds: $c < l$ (low), $l \leq c < h$ (mid-band),
and $c \geq h$ (high). Persistently low scores pull the FSM towards $S0$,
high scores towards $S2$, and sustained mid-band or conflicting signals towards
$S1$ or $S3$. Resource constraints override this logic and force $S4$.
Slow and stopped motion bias $t$ towards larger tiers, while fast motion biases it
towards smaller tiers.

\begin{figure}[t]
\centering
\includegraphics[width=0.8\columnwidth]{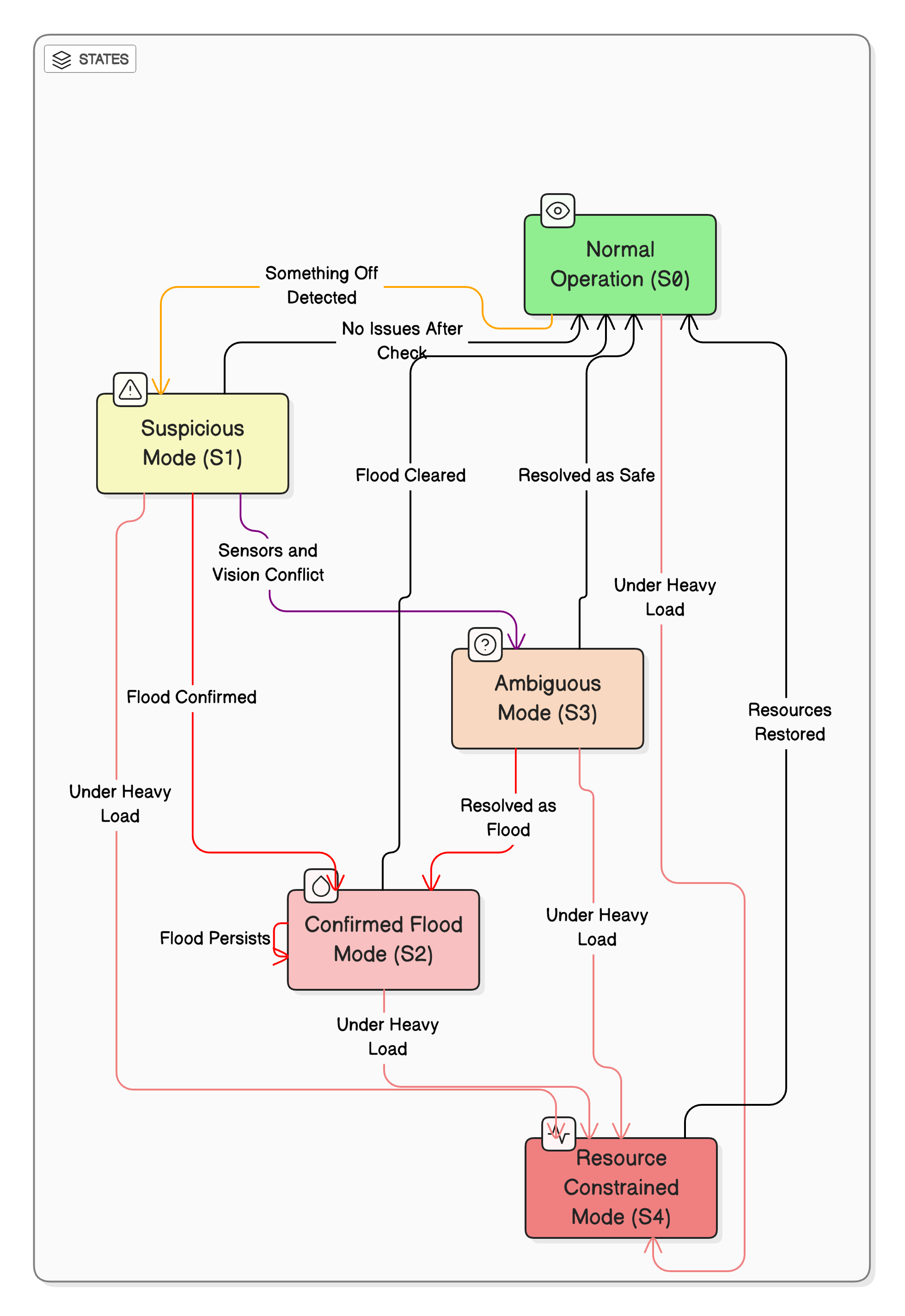}
\caption{High Level Motion-aware finite-state machine}
\label{fig:fsm}
\end{figure}

The intent of each state is:

\begin{itemize}
    \item $S0$: low-power monitoring on the nano tier when scores stay low and signals are consistent.
    \item $S1$: escalate to small/medium tiers when scores linger in the mid-band or become noisy.
    \item $S2$: maintain a flood decision with small/large tiers.
    \item $S3$: resolve disagreements between image and sensor signals by preferring medium/large tiers.
    \item $S4$: degrade to local nano-only inference, then recover towards the previous anchor state once resources stabilise.
\end{itemize}

In all states, nano runs locally on the Processing Pi, while small, medium, and large tiers are offloaded to the Jetson worker by default. 
Figure~\ref{fig:fsm} summarises the states and transitions.

\subsection{Multi-Model YOLO Consensus and Image Score}
\label{sec:decision:consensus}

For each tier, the system uses an ensemble of up to three independently trained YOLOv8 checkpoints (You Only Look Once, a real-time object detector) on distinct datasets (Table~\ref{tab:yolo_models}). When a tier is invoked (locally or on the Jetson), all active models run on the same frame and their detections are aggregated in two steps.

\begin{table}[t]
\centering
\caption{Three-model YOLOv8 ensemble used for consensus aggregation. Datasets: UB Water Detection~\cite{ub_water}, SegWater~\cite{seg_water}, and Ponding-v2~\cite{ponding_v2}.}
\label{tab:yolo_models}
\begin{tabular}{l l c c}
\toprule
Ensemble Slot & Dataset & Images & mAP50 \\
\midrule
Model 1 & UB Water Detection & 2,902 & 92.2\% \\
Model 2 & SegWater & 1,963 & 62.8\% \\
Model 3 & Ponding-v2 & 4,768 & 89.9\% \\
\bottomrule
\end{tabular}
\end{table}

\paragraph*{Multi-model aggregation.}
Raw detections from all models are first flattened and filtered by confidence ($\mathrm{conf} \geq 0.015$). The remaining boxes are then clustered spatially using an IoU threshold of $0.5$: overlapping boxes form a group, while non-overlapping boxes become singleton groups. For each group $G = \{d_i\}$ with boxes $b_i$ and confidences $\mathrm{conf}_i$, the system computes:
\[
C = \sum_i \mathrm{conf}_i,\qquad
\hat{b} = \frac{1}{C} \sum_i \mathrm{conf}_i \, b_i,\qquad
M = |\text{model\_ids}|,
\]
where $C$ is the summed confidence, $\hat{b}$ is the confidence-weighted box, and $M$ is the number of agreeing models. Groups with $C < 0.10$ are discarded. Both overlapping and singleton groups therefore yield \emph{consensus boxes} with confidence $C$ and agreement count $M$.

\paragraph*{Image score.}
Let $A_{\mathrm{img}}$ be the image area and $A_{\mathrm{box},j}$ the area of consensus box~$j$.
Each remaining box contributes
\[
\text{score}_j = C_j \cdot \frac{A_{\mathrm{box},j}}{A_{\mathrm{img}}}
                 \cdot \bigl(1 + 0.2(M_j-1)\bigr)
                 \cdot \frac{3}{N}
\]
where $N$ is the number of models in the ensemble (normalisation) and the term $(1+0.2(M_j-1))$ gives a modest boost to regions seen by multiple models.
The final image score is the sum of all $\text{score}_j$ (or $0$ if no consensus boxes remain).

\subsection{Diurnal Sensor Baselines and Fusion}
\label{sec:decision:sensor}

Sensor fusion uses anomalies relative to diurnal baselines (pre-dawn, midday, evening, night).  
Baselines are maintained as exponentially weighted moving averages (decay constant $\tau \approx 12$\,h) learned only from historically stable ``No Flood'' frames and stored locally.  
At runtime the Processing Pi computes the anomalies $\Delta T$, $\Delta$RH, and $\Delta P$ and applies the physically-motivated rules shown in Table~\ref{tab:sensor_rules} to produce a bounded sensor boost in the range $[-0.08, +0.28]$.  
Both the boost and the raw anomalies are logged per frame.

\begin{table}[t]
\centering
\caption{Sensor boost rules motivated by flash-flood meteorological precursors 
(Doswell et al.~\cite{doswell1996flashflood}) and suppressed runoff under dry antecedent conditions 
(Wasko and Nathan~\cite{wasko2019antecedent}).}
\label{tab:sensor_rules}
\begin{tabular}{l c c}
\toprule
Condition                          & Threshold                            & Boost     \\
\midrule
Humidity rise + cooling            & $\Delta$RH $>$ 15\% and $\Delta T$ $<$ --1.5°C & +0.08 \\
Temperature drop                   & $\Delta T$ $<$ --2.5°C               & +0.04 \\
Pressure fall                      & $\Delta P$ $<$ --5 hPa               & +0.02 \\
Hot \& very dry                    & $\Delta$RH $<$ --20\% and $\Delta T$ $>$ +3.0°C & --0.08   \\
\bottomrule
\end{tabular}
\end{table}

\subsection{Unified Scoring and Decision Provenance}
\label{sec:decision:scoring}
The final \emph{combined score} is the sum of the image score (Section~\ref{sec:decision:consensus}) and the sensor boost.
A frame is classified as No Flood (0) if the combined score is below 0.15, Some Water Detected (1) if in $[0.15, 0.40)$, and Flooded (2) if $\geq 0.40$.
These thresholds (0.15 and 0.40) were empirically selected based on analysis of detections on held-out recorded field videos.
Every frame is accompanied by a deterministic JSON log record that captures complete decision provenance and runtime metadata, enabling full replayability of the system.

\section{Evaluation}
\label{sec:evaluation}
We now evaluate how the proposed architecture behaves under real edge constraints and test the three hypotheses introduced in the introduction: H1 (efficiency from adaptive tiering), H2 (accuracy from multi-model consensus), and H3 (caution from diurnal sensor fusion). The evaluation is based on hardware-in-the-loop replays of recorded field sequences, using deterministic processing and controlled ablations.

\subsection{Experimental Setup and Reproducibility}
We evaluate the system on five field sequences recorded on an off-road trail under consistent lighting and weather. Each is approximately $30\mathrm{s}$ with per-frame ground-truth labels ($0$: No Flood, $1$: Some Water, $2$: Flooded). The sequences cover stopped, slow, and fast motion, with both flooded and dry scenes (Table~\ref{tab:test_sequences}).

\begin{table}[t]
\centering
\caption{Test sequences and motion labels.}
\label{tab:test_sequences}
\begin{tabular}{lccc}
\toprule
Sequence ID & Motion Label & Vehicle Behaviour \\
\midrule
\texttt{slow\_creeping} & slow & Passing flooded section \\
\texttt{fast\_passing} & fast & Toward/away from hazard \\
\texttt{stopped\_water} & stopped & Stationary at water \\
\texttt{stopped\_water\_2} & slow & Around constant flooding \\
\texttt{slow\_no\_water} & slow & No water, fields only \\
\bottomrule
\end{tabular}
\end{table}

To isolate the effect of sensor fusion, we replay identical video sequences under three synthetic sensor regimes:
\begin{itemize}
\item \textbf{Real-wet}: flood-like anomalies ($\Delta T = -3.5\,^\circ\mathrm{C}$,
$\Delta \mathrm{RH} = +18\,\%$, $\Delta P = -8\,\mathrm{hPa}$),
yielding a positive boost of $+0.14$;
\item \textbf{Neutral}: zero anomalies (pure vision baseline, sensor boost $\approx 0.00$);
\item \textbf{Anti-flood}: hot, dry conditions ($\Delta T = +10.0\,^\circ\mathrm{C}$,
$\Delta \mathrm{RH} = -33\,\%$, $\Delta P = +2\,\mathrm{hPa}$),
triggering a negative penalty of $-0.08$.
\end{itemize}
In total, we run $72$ hardware-measured replays across configurations and sensor variants. Replay scripts ensure that all configurations see bit-identical image streams. Reproducibility relies on frozen YOLOv8 weights, per-configuration \texttt{config.py} snapshots, and per-frame JSON decision logs on the Processing~Pi; all processing is deterministic, so replays from the same input stream yield identical decisions. The replication package, comprising the Processing/Jetson and Gathering Pi codebases used to produce the ablation results in Table~\ref{tab:overall_performance}, is submitted with this paper and will be made publicly available for readers.\footnote{Processing/Jetson: \url{https://github.com/Oliver1703dk/flood_detection_system}; Gathering: \url{https://github.com/Oliver1703dk/edge_data_collector}}

Internally the system uses three classes, with {\em Some Water} acting as a watch state. For evaluation we use a binary scheme: {\em No Flood} and {\em Some Water} are merged into a single non-flood class, while {\em Flooded} remains as {\em Flood}. This focuses metrics on detecting confirmed floods when it matters and treats the watch state as non-alert.

\subsection{Ablation Configurations}
We evaluate ten ablation configurations to isolate the architectural tactics: FSM orchestration, multi-model consensus, sensor fusion, and offloading policies. Configurations vary model tiers (nano/small/medium/large), FSM enablement, ensemble size (one vs.\ three models), sensor fusion, and Jetson usage. All image scores use the same normalisation factor ($3.0 / \text{num\_models}$) so that unified thresholds and FSM behaviour remain comparable across ensembles.
Table~\ref{tab:ablations} summarises the configurations. The production configuration (ID~4) enables all tactics: FSM-guided tiering, three-model consensus, diurnal sensor fusion, and adaptive offloading. Baselines such as \texttt{static\_nano} (1b) represent realistic local-only operation, while \texttt{always\_offload\_medium} (6) approximates a naive ``always use the heavy model'' policy.

\begin{table*}[t]
\centering
\caption{Ablation configurations.}
\label{tab:ablations}
\begin{tabular}{lcccccc}
\toprule
ID & Models/Tiers & FSM & Sensor Fusion & Offload Policy & Jetson & Purpose \\
\midrule
1 & 1$\times$ small & Off & Off & No & No & Static single-model upper bound \\
1b & 1$\times$ nano & Off & Off & No & No & Realistic static local baseline \\
2 & 3$\times$ n/s/m/l & On & Off & Adaptive & Yes & Vision-only FSM with consensus \\
2b & 1$\times$ n/s/m/l & On & Off & Adaptive & Yes & Consensus ablation in vision FSM \\
3 & 3$\times$ nano & On & On & No & No & Full system without Jetson \\
3b & 1$\times$ nano & On & On & No & No & Consensus ablation local-only \\
4 & 3$\times$ n/s/m/l & On & On & Adaptive & Yes & Production system \\
4b & 1$\times$ n/s/m/l & On & On & Adaptive & Yes & Consensus ablation in production \\
5 & 3$\times$ n/s/m/l & On & On & Force medium (fast) & Yes & Fast-motion safety variant \\
6 & 1$\times$ medium & Off & Off & Always & Yes & Naive always-offload baseline \\
\bottomrule
\end{tabular}
\end{table*}

\subsection{Metrics and Analysis Method}
We report both classification quality and system-level behaviour using the binary scheme:
\begin{itemize}
\item \textbf{Classification metrics}: binary macro F1, balanced accuracy, flood precision and recall (with emphasis on flood recall for safety).
\item \textbf{Latency and energy}: end-to-end latency percentiles (p99) and total energy per video (J per $30\,\mathrm{s}$), computed from hardware power measurements.
\item \textbf{Stability and allocation}: label oscillation count (unnecessary label changes), FSM state traces, tier usage distributions, and temporal coverage (percentage of frames that receive a decision).
\end{itemize}
These metrics support the three hypotheses as follows:
\begin{itemize}
\item \textbf{H1 (Efficiency)}: total energy, heavy-tier usage, and temporal coverage;
\item \textbf{H2 (Accuracy)}: binary macro F1 and flood recall vs.\ ensemble size;
\item \textbf{H3 (Caution)}: sensor-variant comparisons of watch/flood decision share, temporal coverage, and flood recall under different environmental regimes.
\end{itemize}
Aggregated results are computed across the subset of sequences where all compared configurations were run, to ensure fair comparisons. We now present results grouped by hypothesis.

\subsubsection{H1: Efficiency from Adaptive Tiering}
\begin{figure}[t]
\centering
\includegraphics[width=\columnwidth]{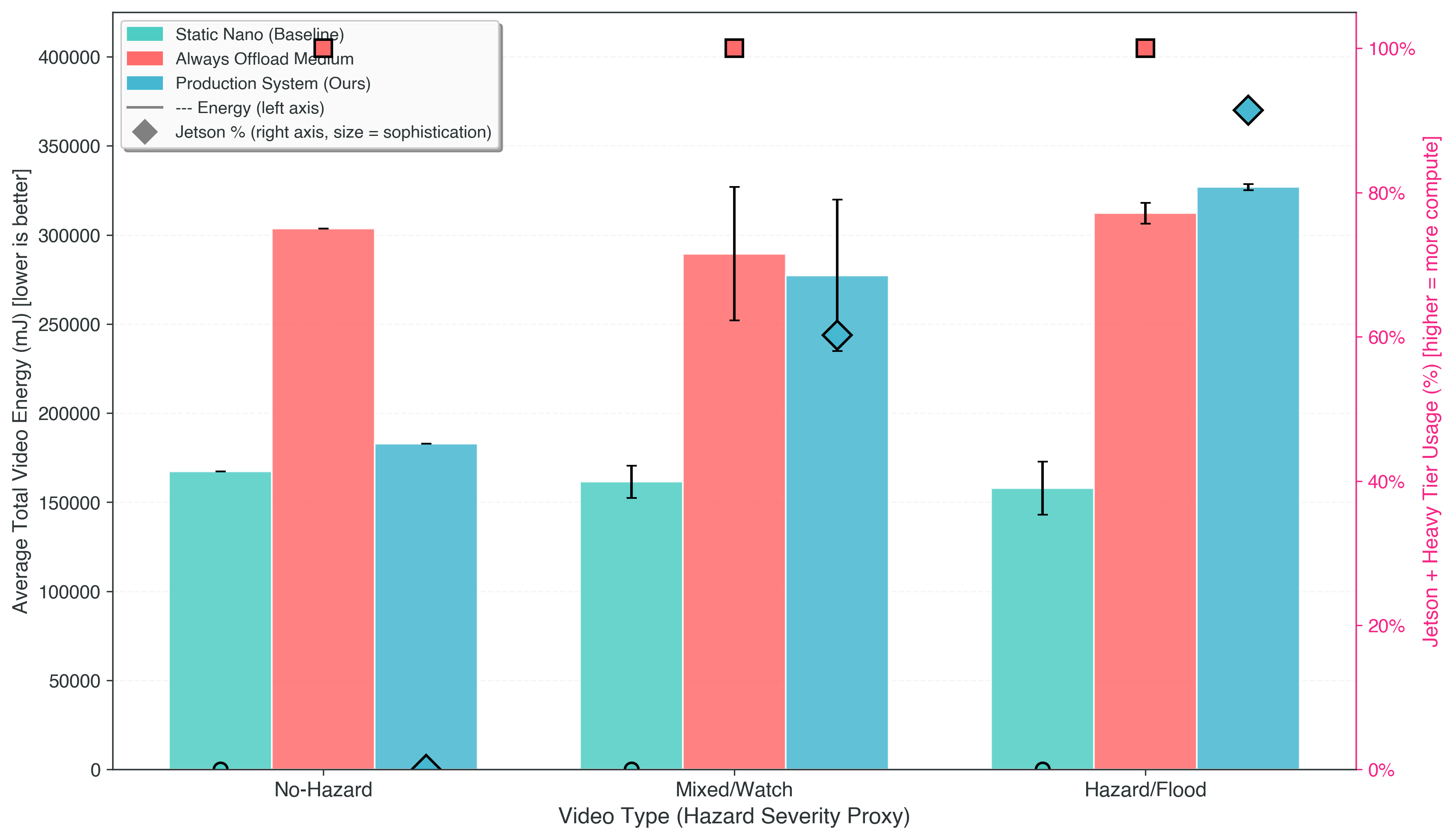}
\caption{Adaptive behaviour across hazard severity: grouped bars for energy (left axis) and heavy-tier usage (right axis) by regime for baselines and production models.}
\label{fig:adaptive}
\end{figure}

H1 states that adaptive tiering and offloading should reduce energy usage compared to naive heavy-model policies, while maintaining comparable classification quality. We therefore compare the production configuration (\texttt{production}) against static local baselines and the naive \texttt{always\_offload\_medium} configuration.
Figure~\ref{fig:adaptive} summarises adaptive behaviour across hazard regimes by plotting total energy and heavy-tier usage for \texttt{static\_nano}, \texttt{always\_offload\_medium}, and \texttt{production}. On dry sequences such as \texttt{slow\_no\_water}, the production system closely matches \texttt{static\_nano} in energy. On mixed and flooded sequences, it progressively increases heavy-tier utilisation and energy to maintain recall, but still uses substantially fewer heavy-tier frames than \texttt{always\_offload\_medium}.

Table~\ref{tab:overall_performance} aggregates performance across the full test set. Compared to \texttt{always\_offload\_medium}, the production configuration reduces total energy (from $299.4\mathrm{J}$ to $278.0\mathrm{J}$) while achieving higher binary macro F1 ($0.791$ vs.\ $0.816$). Relative to \texttt{static\_nano}, production consumes more energy ($161.3\mathrm{J}$ vs.\ $278.0\mathrm{J}$) but improves binary macro F1 from $0.627$ to $0.816$ and flood recall from $0.401$ to $0.812$. Overall, these results support H1: the FSM-guided tiering policy focuses heavy tiers on hazard segments rather than expending energy uniformly.

\begin{table*}[t]
\centering
\caption{Overall performance across the full test set.}
\label{tab:overall_performance}
\begin{tabular}{llccccc}
\toprule
ID & Configuration & Total Energy (J) & Macro F1 (2-class) & Balanced Accuracy & p99 Lat. (ms) & Temporal Coverage \\
\midrule
1   & \texttt{static\_small}          & 171.7 & 0.755 & 0.733 & 4913.4 & 0.412 \\
1b  & \texttt{static\_nano}           & 161.3 & 0.627 & 0.629 & 2552.3 & 0.470 \\
2   & \texttt{vision\_fsm\_multi}     & 262.0 & 0.785 & 0.789 & 2587.6 & 0.819 \\
2b  & \texttt{vision\_fsm\_single}    & 265.5 & 0.783 & 0.785 & 1137.2 & 0.935 \\
3   & \texttt{full\_local\_multi}     & 167.6 & 0.740 & 0.726 & 2638.0 & 0.462 \\
3b  & \texttt{full\_local\_single}    & 167.3 & 0.638 & 0.639 & 1209.8 & 0.575 \\
4   & \texttt{production (ours)} \textsuperscript{$\star$} 
                                     & 278.0 & 0.816 & 0.816 & 2573.4 & 0.841 \\
4b  & \texttt{production\_single}     & 283.8 & 0.757 & 0.767 & 1181.6 & 0.911 \\
5   & \texttt{fast\_force\_jetson}    & 272.6 & 0.807 & 0.804 & 2086.1 & 0.847 \\
6   & \texttt{always\_offload\_medium}& 299.4 & 0.791 & 0.800 & 375.9  & 0.936 \\
\bottomrule
\end{tabular}
\end{table*}

\subsubsection{H2: Accuracy from Multi-Model Consensus}
H2 states that multi-model consensus improves accuracy and calibration.  
We compare the three paired configurations that differ only in ensemble size:  
\texttt{vision\_fsm} (2 vs 2b), \texttt{full\_local} (3 vs 3b), and \texttt{production} (4 vs 4b).

Table~\ref{tab:overall_performance} shows that switching to three-model consensus increases binary macro F1 by 6--10 percentage points (average +7.9 pp) and balanced accuracy by a similar amount, while flood recall remains essentially unchanged.  
Energy consumption stays virtually identical (multi-model configurations use on average $\sim$1\% less energy than their single-model counterparts).  
The primary trade-off is tail latency, which roughly doubles (p99 $\sim$ 2.5 s vs $\sim$ 1.2 s) due to executing three models instead of one.

These results confirm H2: multi-model consensus consistently improves accuracy.

\subsubsection{H3: Robustness from Sensor Fusion}
H3 states that diurnal sensor fusion should modulate system caution across environmental regimes: flood-like anomalies should make the system more vigilant and compute-hungry, while hot, dry anomalies should stabilise decisions, without substantially degrading flood recall.

To isolate sensor effects, we run the production configuration on \texttt{slow\_creeping} with the three synthetic sensor variants described earlier. All runs use bit-identical video, so any differences arise solely from the sensor fusion branch. Table~\ref{tab:sensor_fusion_effect} summarises temporal coverage and watch/flood recall.

\begin{table}[t]
\centering
\caption{Effect of sensor fusion.}
\label{tab:sensor_fusion_effect}
\begin{tabular}{lcccc}
\toprule
Variant    & Frames & Temp. Cov. & Flood R. & Watch R. \\
\midrule
Neutral    & 21 & 0.667 & 0.917 & 0.286 \\
Real-wet   & 29 & 0.917 & 0.923 & 0.833 \\
Anti-flood & 20 & 0.667 & 1.000 & 0.167 \\
\bottomrule
\end{tabular}
\end{table}

Relative to the Neutral (vision-only) baseline, the Real-wet variant processes more frames and increases temporal coverage (from $0.667$ to $0.917$) and watch recall (from $0.286$ to $0.833$), while keeping flood recall high ($0.917$ to $0.923$). The Anti-flood variant moves in the opposite direction: temporal coverage returns to $0.667$, watch recall drops to $0.167$, and flood recall reaches $1.0$. These results support H3: sensor branch reshapes caution and temporal coverage.

\subsubsection{Summary of Findings}
Across hypotheses, the evaluation shows that:
\begin{itemize}
\item FSM-guided tiering concentrates heavy tiers on hazard segments, reducing energy relative to naive always-offload policies while maintaining or improving binary macro F1 (H1);
\item multi-model consensus improves binary macro F1 by around $0.07$ absolute compared to single-model variants at similar or lower energy (H2); and
\item diurnal sensor fusion modulates caution, increasing temporal coverage and watch decisions in wet regimes and yielding conservative behaviour in hot, dry regimes, without changing the visual input (H3).
\end{itemize}
The next section discusses these trade-offs in more detail and relates them back to architectural decisions such as FSM design and sensor weighting.

\subsection{Discussion}
%\label{sec:discussion}

This subsection discusses how the evaluation results for H1--H3 inform the architecture and its limitations.

\paragraph*{Hypothesis Summary}
As previously detailed, the results support H1--H3: FSM-guided tiering reduces energy vs.\ naive policies while improving F1 (Table~\ref{tab:overall_performance}); multi-model consensus boosts accuracy by $\sim$8 pp at near-identical energy (H2); and sensor fusion modulates caution across regimes without harming recall (Table~\ref{tab:sensor_fusion_effect}, H3).

\paragraph*{Architectural Implications and Limitations}
These tactics form a coherent trade-off space: FSM tiering focuses compute on hazards, consensus enhances reliability, and fusion adapts caution environmentally. The architecture tolerates multi-second p99 latencies and partial temporal coverage, as bounded queues and offload timeouts favour responsiveness over processing every frame. This is suitable for the evaluated off-road flood-warning setting but would constrain use in higher-speed or tighter-control domains. Additional limitations include the limited sequence diversity (no night or heavy rain), platform-specific results (RPi+Jetson), and synthetic sensor variants for H3, motivating the extensions in Section~\ref{sec:conclusion}.

\subsection{Threats to Validity}
\textbf{Internal validity.} Measurement noise is reduced by using p99 latency, outlier removal (IQR filtering), fixed clock speeds, and deterministic replay of bit-identical streams.

\textbf{External validity.} Our sequences represent typical agricultural conditions (dry, watch, flooded) under moderate weather and lighting, but do not cover extreme cases (heavy rain, snow, night, dust storms). Results were obtained on Raspberry-Pi-class boards with an optional Jetson; other platforms may exhibit different energy–latency trade-offs.

\textbf{Construct validity.} We evaluate binary flood detection (merging “No Flood” and “Some Water”), which prioritises safety-critical alerts but may undervalue the predictive role of the intermediate “watch” state. Long-term effects such as sensor drift and driver trust are approximated via diurnal baselines and label stability metrics rather than fleet-scale deployment.

\section{Conclusion and Future Work}
\label{sec:conclusion}

This paper presented a deployed edge architecture for standing-water detection on off-road vehicles that combines FSM-guided tiering, multi-model YOLO consensus, and diurnal sensor fusion in a single stateful control plane. The architecture separates concerns across Gathering, Processing, and Jetson nodes, but centralises long-lived architectural state and decision logic on the Processing Pi with unified scoring and schema-stable messages. Deterministic per-frame logs and replayable storage layouts make the behaviour executable and auditable, allowing architectural tactics to be evaluated under realistic hardware constraints. 
The evaluation across ten ablation configurations and sensor variants supports the three hypotheses that structure the work. For H1, FSM-guided tiering and motion-aware offloading concentrate heavy tiers on hazard segments instead of expending compute uniformly, improving binary macro F1 and temporal coverage relative to static local baselines while using less energy than a naive always-heavy policy. For H2, multi-model consensus turns additional compute into more reliable decisions, consistently increasing binary macro F1 and balanced accuracy at essentially unchanged energy, with the main trade-off being higher tail latency. For H3, diurnal sensor fusion reshapes caution: flood-like anomalies increase temporal coverage and watch decisions, while hot, dry anomalies promote more conservative behaviour, without harming flood recall on the evaluated sequence. Together, these tactics define a coherent trade-off space in which accuracy, caution, latency, and energy can be tuned at the architectural level rather than only via model training.
Several aspects remain for future work. On the sensing and fusion side, we plan to extend diurnal baselines to support online adaptation under long-term drift and to incorporate additional modalities such as IMU, radar, or soil moisture sensors. On the control side, an important direction is to explore adaptive or learned FSM policies that adjust thresholds and escalation rules based on deployment feedback, while preserving interpretability and replayability. From a validation perspective, we aim to broaden evaluation to more diverse weather, lighting, and terrain conditions, including night-time and extreme events, and to larger fleets and heterogeneous hardware platforms. Finally, we see potential in complementing the current numeric provenance with explainability mechanisms, e.g., lightweight vision–language models or templated narratives, that can expose the system’s hazard assessments and resource-allocation decisions to human operators.

\balance
\bibliographystyle{IEEEtran}
\bibliography{references}
\end{document}